\documentclass[sigconf]{acmart}
\AtBeginDocument{%
  }

\setcopyright{acmlicensed}
\copyrightyear{2026}
\acmYear{2026}
\acmDOI{XXXXXXX.XXXXXXX}

\renewcommand\footnotetextcopyrightpermission[1]{}

\acmConference[XX]{Make sure to enter the correct
  conference title from your rights confirmation emai}{2026}{XX, XXX}
\acmISBN{978-1-4503-XXXX-X/2018/06}

\usepackage[utf8]{inputenc} 
\usepackage[T1]{fontenc}    
\usepackage{hyperref}       
\usepackage{url}            
\usepackage{booktabs}       
\usepackage{amsfonts}       
\usepackage{nicefrac}       
\usepackage{microtype}      
\usepackage{xcolor}         
\usepackage{amsmath} 
\usepackage{tabularx} 
\usepackage{booktabs} 
\usepackage{wrapfig}
\usepackage{adjustbox}
\usepackage{booktabs}    
\usepackage{multirow}    
\usepackage{rotating}    
\usepackage{makecell}    
\usepackage{graphicx}    
\usepackage{xcolor}      
\usepackage{colortbl}    
\usepackage{subcaption} 
\usepackage{wrapfig}  
\usepackage{graphicx} 
\usepackage{booktabs}  
\usepackage{algorithm}
\usepackage{algpseudocode}
\usepackage{amsmath}
\usepackage{enumitem}
\usepackage[normalem]{ulem}

\begin{document}

\title{GraphCogent: Mitigating LLMs' Working Memory Constraints via Multi-Agent Collaboration in Complex Graph Understanding}

\pdfstringdefDisableCommands{%
  \def\textsuperscript#1{#1}%
  \def\\{ }%
}

\author{
Rongzheng Wang\textsuperscript{1},
Shuang Liang\textsuperscript{1*},
Qizhi Chen\textsuperscript{1},
Yihong Huang\textsuperscript{1},
Muquan Li\textsuperscript{1},\\
{}Yizhuo Ma\textsuperscript{1},
Dongyang Zhang\textsuperscript{1},
Ke Qin\textsuperscript{1},
Man-Fai Leung\textsuperscript{2}\\[0.6ex]
\textsuperscript{1} University of Electronic Science and Technology of China\\
\textsuperscript{2} Anglia Ruskin University\\[0.6ex]
\texttt{wangrongzheng@std.uestc.edu.cn}\quad
\texttt{shuangliang@uestc.edu.cn}
}

\renewcommand{\shortauthors}{Wang et al.}

\begin{abstract}
  Large language models (LLMs) show promising performance on small-scale graph reasoning tasks but fail when handling real-world graphs with complex queries. This phenomenon arises from LLMs' working memory constraints, which result in their inability to retain long-range graph topology over extended contexts while sustaining coherent multi-step reasoning. However, real-world graphs are often structurally complex, such as Web, Transportation, Social, and Citation networks. To address these limitations, we propose \textbf{GraphCogent}, a collaborative agent framework inspired by human \textit{Working Memory Model} that decomposes graph reasoning into specialized cognitive processes: sense, buffer, and execute. The framework consists of three modules: Sensory Module standardizes diverse graph text representations via subgraph sampling, Buffer Module integrates and indexes graph data across multiple formats, and Execution Module combines tool calling and tool creation for efficient reasoning. We also introduce Graph4real, a comprehensive benchmark that contains four domains of real-world graphs (Web, Transportation, Social, and Citation) to evaluate LLMs' graph reasoning capabilities. Our Graph4real covers 21 different graph reasoning tasks, categorized into three types (Structural Querying, Algorithmic Reasoning, and Predictive Modeling tasks), with graph scales up to 10 times larger than existing benchmarks. Experiments show that Llama3.1-8B based GraphCogent achieves a 50\% improvement over massive-scale LLMs like DeepSeek-R1 (671B). Compared to state-of-the-art agent-based baseline, our framework outperforms by 20\% in accuracy while reducing token usage by 80\% for in-toolset tasks and 30\% for out-toolset tasks. Code will be available after review.
\end{abstract}




\maketitle

\section{Introduction}
\label{secintro}

Large language models (LLMs)~\cite{DBLP:journals/corr/abs-2303-08774} have demonstrated remarkable cognitive capabilities in natural language processing. However, when querying the connection path between two webpages in a Web graph, state-of-the-art LLMs like DeepSeek-R1~\cite{DBLP:journals/corr/abs-2501-12948}, GPT-o3~\cite{DBLP:journals/corr/abs-2303-08774}, and Gemini-2.5 pro~\cite{DBLP:journals/corr/abs-2312-11805} return incorrect navigation routes in 9 out of 10 cases. This failure reveals that current LLMs remain limited in handling large real-world graph reasoning tasks.

\begin{figure}
\centering
\includegraphics[width=\linewidth]{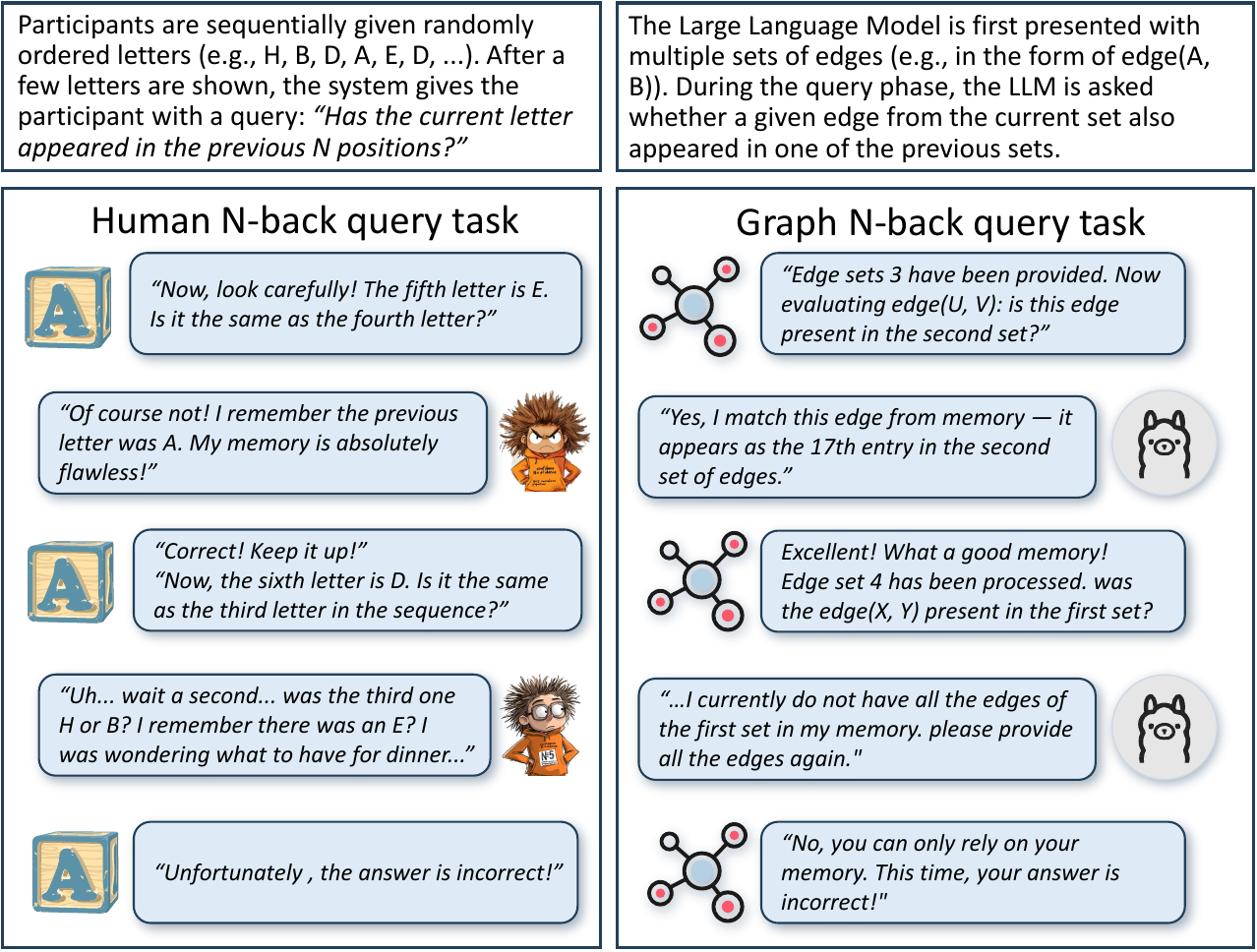}
\caption{
    Graph N-back Query Task: A graph is split into 50-edge subsets $E_t$. At turn $t+N$, the LLM verifies edge existence in $E_t$. Experimental results are in Section~\ref{Section 3.1}.
}
\label{Chapter1_intro}
\vspace{-2em}
\end{figure}

Researchers have explored various methods to address the constraints of LLMs' inability to perform graph reasoning. \textit{Text-based} methods~\cite{DBLP:journals/corr/abs-2310-04560, DBLP:journals/corr/abs-2403-04483, DBLP:journals/corr/abs-2402-16029} leverage Chain-of-Thought (CoT)~\cite{DBLP:conf/nips/Wei0SBIXCLZ22} reasoning but struggle with the dynamic programming requirements of graph algorithms (e.g., Bellman-Ford~\cite{bellman1958routing}), resulting in cascading errors in complex tasks. \textit{Tool-based} methods~\cite{DBLP:journals/corr/abs-2304-11116, DBLP:conf/kdd/Wang0CZQ25} rely on external tools for computation, requiring predefined tools and rigid input formats (e.g., preprocessed files). These limitations hinder their ability to process diverse graph text representations, such as adjacency lists, symbolic notations, and linguistic descriptions~\cite{DBLP:journals/corr/abs-2310-04560}, thereby reducing their adaptability to real-world scenarios. \textit{Agent-based} methods~\cite{DBLP:journals/corr/abs-2410-18032, zhang2024gcoder} employ multi-agent collaboration, often decomposing graph tasks into sequential stages via naive instruction prompts. However, these approaches are fundamentally limited by the complexity of code construction and the inherent memory constraints of LLMs. As graph scales increase and tasks involve real-world graph problems, agent-based methods perform poorly when handling graph data comprehension and task execution.

This limitation stems from LLMs' working memory constraints that restrict simultaneous processing of complex graph topology and multi-step reasoning. A common mitigation approach involves partitioning graphs into smaller units for sequential processing~\cite{DBLP:conf/emnlp/JiangZDYZW23}, mirroring how humans decompose complex problems into manageable information chunks~\cite{miller1956magical}. Classic human working memory capacity assessments N-back test~\cite{Kirchner_2006, Klatzky_Giudice_Marston_Tietz_Golledge_Loomis_2008, Amon_Bertenthal_2018} (exampled in Fig.~\ref{Chapter1_intro}, left) reveals a rapid cognitive decline when humans are required to retain more than three items. Inspired by the N-back test, we design an analogous LLMs' Graph N-back test. As exampled in the right of the Fig.~\ref{Chapter1_intro}, by the 3rd processing turn, Llama3.1-8B~\cite{DBLP:journals/corr/abs-2302-13971} exhibits significant topology forgetting, detailed quantitative result is analyzed in Section~\ref{Section 3.1}. This demonstrates that LLM graph reasoning similarly suffers from working memory constraints, mirroring human cognitive limitations.

Human brain addresses these constraints through a specialized working memory model for sensory processing, information buffering, and task execution~\cite{DBLP:journals/jocn/BaddeleyJV11}. Facing a complex task, humans rely on an external sensory system to perceive stimuli, an episodic buffer to integrate and store information, and a central executive to coordinate cognition and processing. This absent architecture in monolithic LLMs reveals their key graph reasoning bottlenecks: 
(1) \textit{Diverse Text Representations}: The sensory limitation in processing diverse graph text representations leads to inconsistent comprehension, especially for mainstream open-source models (e.g., Llama~\cite{DBLP:journals/corr/abs-2302-13971}, Qwen~\cite{yang2024qwen2}, and GLM~\cite{glm2024chatglm}), thus impairing reasoning accuracy; (2) \textit{Overload Graph Scale}: Lacking a buffer mechanism, LLMs' limited input context windows hinders long-range dependency capture, causing global information loss and cascading reasoning errors; (3) \textit{Code Execution Fragility}: The executive dysfunction results in unreliable code generation and inefficient algorithmic implementation.

Inspired by the cognitive architecture of human working memory model, we propose \textbf{GraphCogent}, a \textbf{Graph} \textbf{Co}llaboration A\textbf{gent}ic framework designed to mitigate LLMs’ working memory constraints in graph reasoning. Our framework consists of three modules: \textbf{Sensory Module} employs a sensory agent to sample subgraph and transform unstructured graph text representations into standardized adjacency list. \textbf{Buffer Module} integrates graph data from the Sensory Module and establishes data indices for diverse data formats (including NumPy, PyG, and NetworkX) based on Buffer Agent. \textbf{Execution Module} synergistically combines tool calling and with on-demand tool creation. Naive graph reasoning tasks are handled by a Execution Agent through a pre-built common toolset, while complex tasks leverage a tool creator that produces modular, task-specific components and backfills new tool requests with the Buffer’s preprocessed data, thereby avoiding error-prone full-code generation.

To evaluate LLM's generalization capability in graph reasoning, we construct a more challenging Graph4real benchmark dataset, which collects graph data from four real-world domains: Web~\cite{DBLP:conf/nips/McAuleyL12}, Social~\cite{DBLP:conf/nips/McAuleyL12}, Transportation~\cite{DBLP:journals/spm/ShumanNFOV13}, and Citation~\cite{DBLP:conf/aaai/CravenFMMNS98}.
Our Graph4real covers 21 different graph reasoning tasks, categorized into three types (Structural Querying, Algorithmic Reasoning, and Predictive Modeling tasks), with graph scales that are 10 times larger than existing benchmarks (e.g., NLGraph~\cite{DBLP:conf/nips/WangFHTHT23}, GraphWiz~\cite{DBLP:journals/corr/abs-2402-16029}). Experimental results demonstrate that our method achieves state-of-the-art performance with an average accuracy of 98.5\% based on the open-source Llama3.1-8B, representing a 20\% improvement over other agent-based approaches. Cross-dataset validation demonstrates robust generalization, maintaining over 90\% accuracy across diverse public benchmarks. The combining tool calling and tool creation strategy delivers dual optimization: it reduces token consumption by 80\% for tasks covered by our toolset while still achieving 30\% token savings for out-toolset tasks compared to the conventional agent-based method.

Our contributions are summarized as follows: (1) We propose GraphCogent, a novel agent collaboration framework addressing LLMs' working memory constraints through its sensory-buffer-execution architecture; (2) We design a reasoning mechanism that combines tool calling and tool creation with a \textit{Thinking Policy Initialization} to stabilize early-stage decision-making and a tailored \textit{Capability-Margin Preference Optimization (CMPO)} objective to sharpen tool discrimination, enabling efficient handling of diverse graph tasks; (3) We construct Graph4real, a comprehensive benchmark featuring real-world graphs that covers 21 different graph reasoning tasks, categorized into three types (Structural Querying, Algorithmic Reasoning, and Predictive Modeling tasks), with graph scales that are 10 times larger than existing benchmarks.

\section{Related work}
\textbf{LLMs for Graph Reasoning Tasks.}
Existing graph reasoning methods adopt three paradigms: \textit{Text-based}~\cite{DBLP:journals/corr/abs-2310-04560, DBLP:journals/corr/abs-2403-04483, DBLP:journals/corr/abs-2402-16029, DBLP:journals/corr/abs-2305-15066, DBLP:conf/acl/WangWH00M24} approaches leverage chain-of-thought prompting to decompose graph algorithms stepwise, but suffer error accumulation in multi-step reasoning. \textit{Tool-based}~\cite{DBLP:journals/corr/abs-2304-11116, DBLP:conf/kdd/Wang0CZQ25, DBLP:journals/corr/abs-2303-08774} methods offload computations to external solvers (e.g., NetworkX), yet rigid input requirements limit adaptability to diverse text representations. \textit{Agent-based}~\cite{DBLP:journals/corr/abs-2410-18032, zhang2024gcoder} frameworks attempt task decomposition through multi-agent collaboration, but face scalability bottlenecks when handling large graphs due to code generation complexity and memory overload.

\textbf{Working Memory Constraints in LLMs.} Like humans facing working memory constraints in complex tasks~\cite{doi:10.1073/pnas.0801268105, 10.1007/bf03392069}, LLMs show similar capacity constraints. Cognitive studies show humans typically retain only few information units simultaneously, with performance decaying sharply in N-back tests (where N = 3)~\cite{Kirchner_2006, Klatzky_Giudice_Marston_Tietz_Golledge_Loomis_2008, Amon_Bertenthal_2018, Jaeggi_Buschkuehl_Perrig_Meier_2010}. LLMs mirror this pattern, exhibiting accuracy declines on 3-step dependencies~\cite{DBLP:books/el/74/BaddeleyH74}. This working memory overload manifests in graph reasoning as excessive graph scale and complex reasoning tasks, thus hindering effective processing of intricate graph structures and multi-step reasoning.

\textbf{Graph Reasoning Benchmarks for LLMs.} NLGraph~\cite{DBLP:conf/nips/WangFHTHT23}, as the pioneering benchmark, demonstrates LLMs' competence on small-scale graphs (less than 40 nodes) with basic reasoning tasks. GraphWiz~\cite{DBLP:journals/corr/abs-2402-16029} later expands to 100 nodes with more complex tasks. However, these and other benchmarks~\cite{DBLP:journals/corr/abs-2403-04483, DBLP:conf/acl/WangWH00M24, DBLP:journals/corr/abs-2310-04560, DBLP:conf/kdd/Wang0CZQ25} remain constrained by fixed-size graphs within context windows and templated reasoning tasks, failing to match real-world reasoning demands where both scale and complexity exceed current test conditions.

\section{Graph4real Construction}
\label{Graph4realConstruction}
Current LLMs' graph reasoning benchmarks face three key limitations: (1) Limited scale, primarily using randomly generated graphs that lack real-world topological characteristics; (2) Simple textual representations, over-relying on preprocessed files or adjacency lists while ignoring domain-specific semantic descriptions; (3) Artificial task formulations, where queries directly specify algorithms with only a single sentence rather than mirroring real-world reasoning scenarios requiring intent interpretation.

\textbf{Real-World Graph Scaling.} To establish a comprehensive benchmark for evaluating LLMs on real-world graph reasoning tasks, we curate a diverse set of graphs from four domains: Web (Google Web Graph~\cite{DBLP:conf/nips/McAuleyL12}), Social (SNAP Social Circles~\cite{DBLP:conf/nips/McAuleyL12}), Transportation (PeMS~\cite{DBLP:journals/spm/ShumanNFOV13}), and Citation graphs (Cora~\cite{DBLP:conf/aaai/CravenFMMNS98}). Unlike prior benchmarks~\cite{DBLP:conf/nips/WangFHTHT23, DBLP:journals/corr/abs-2403-04483} that rely on randomly generated graphs, our dataset is constructed from real-world sources to ensure realistic topological properties and domain-specific characteristics. We employ a biased random walk sampling strategy to construct graphs at three scales (40, 100, and 1000 nodes), ensuring both effective evaluation of LLM-based methods through smaller-scale graphs and construct large-scale graphs to challenge existing methods.

\textbf{Various Graph Text Representation.} We formalize three graph text representation formats to evaluate LLMs' graph comprehension and reasoning capabilities: (1) Adjacency list ([0,1],[0,2],...); (2) Symbolic notation (0→1, 2→3); (3) Linguistic descriptions using domain-specific predicates (Linked/Followed/Connected/Cited). These representations preserve various semantics, enabling assessment of LLMs' ability to process both formal graph formats and domain-specific descriptions. 

\begin{figure}[t]
  \centering
  \includegraphics[width=\linewidth,keepaspectratio]{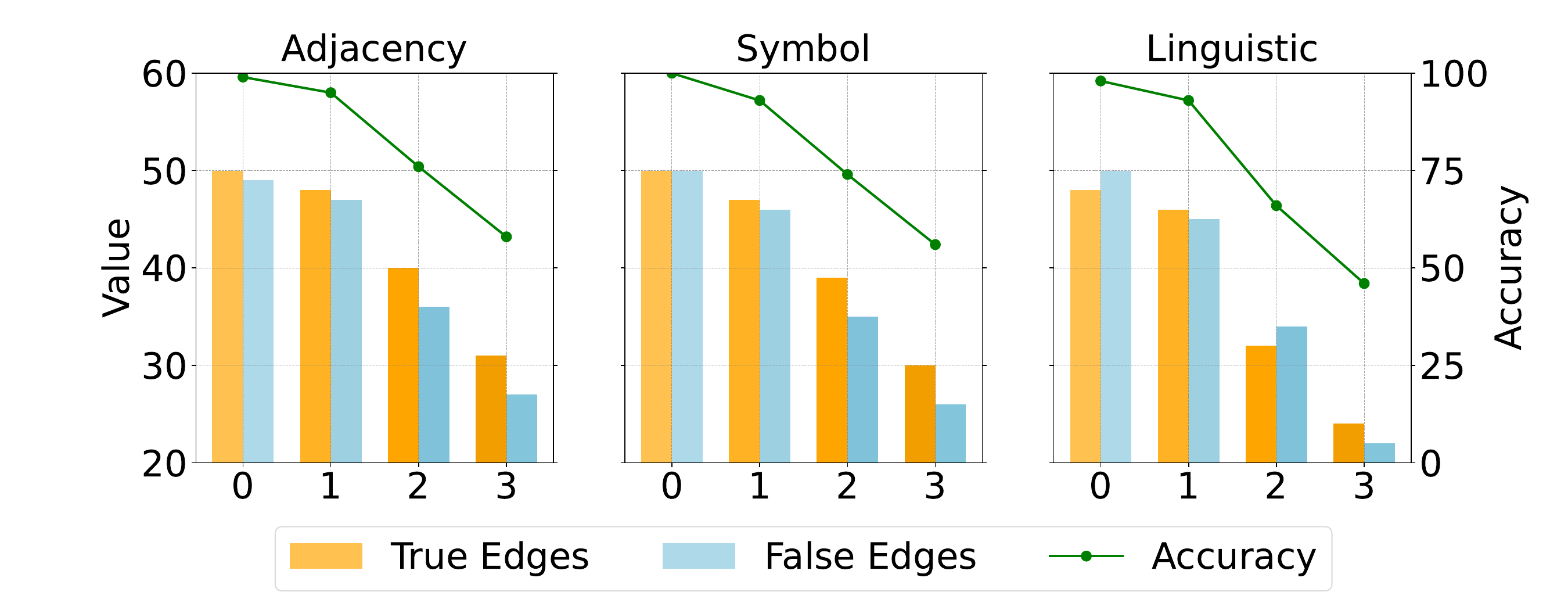}
  \caption{Graph N-back Query. Accuracy (right y-axis) measures memory retention across dialogues, computed as sum of True edges (correctly identified existing edges) and False edges (correctly rejected non-existent edges) (left y-axis).}
  \label{Chapter3_Nback}
\includegraphics[width=\linewidth,keepaspectratio]{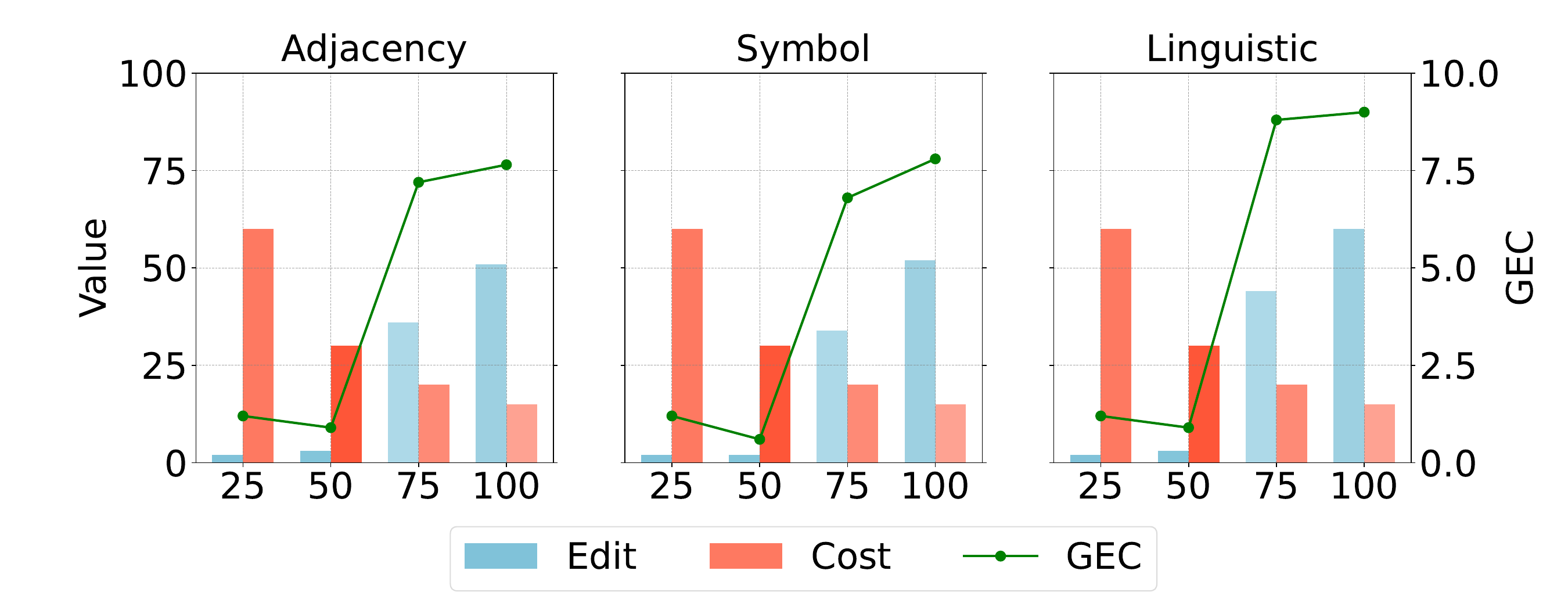}
  \caption{Graph Efficiency Coefficient (right y-axis) across sampling granularity. Edit shows graph Edit distance (left y-axis), and Cost (left y-axis, scaled ×100) counts total LLM interactions. Lower GEC indicates better performance.}
  \label{Chapter3_Representation}
\end{figure}

\textbf{Intent-Driven Task Design.} A total number of 21 tasks are designed and categorized into three classes: (1) \textit{Graph Structural Querying Tasks}: Focus on fundamental graph properties, such as edge existence or node count, testing basic graph structural comprehension. (2) \textit{Graph Algorithmic Reasoning Tasks}: Require reasoning based on classical algorithms, such as shortest path and maximum flow, evaluating algorithmic proficiency. (3) \textit{Graph Predictive Modeling Tasks}: Involve neural network-based predictions, such as node classification and link prediction, assessing predictive capabilities.

For task generation, we collect five real-world scenarios for each of the four domains (e.g., travel planning and logistics optimization for transportation) and craft 20 prompt templates to construct contextually grounded questions. Using DeepSeek-R1~\cite{DBLP:journals/corr/abs-2501-12948}, we generate graph reasoning tasks based on templates with specific scenarios, ensuring both diversity and relevance. To validate task quality, we employ a dual evaluation approach: DeepSeek-R1 for automated assessment and human for manual review. This process produces a dataset with 4200 questions distributed across the 21 tasks. The detailed task definitions, dataset statistics, and prompt templates are provided in our released anonymous github code.

\section{GraphCogent}
\begin{figure*}
  \centering
  \includegraphics[width=\linewidth]{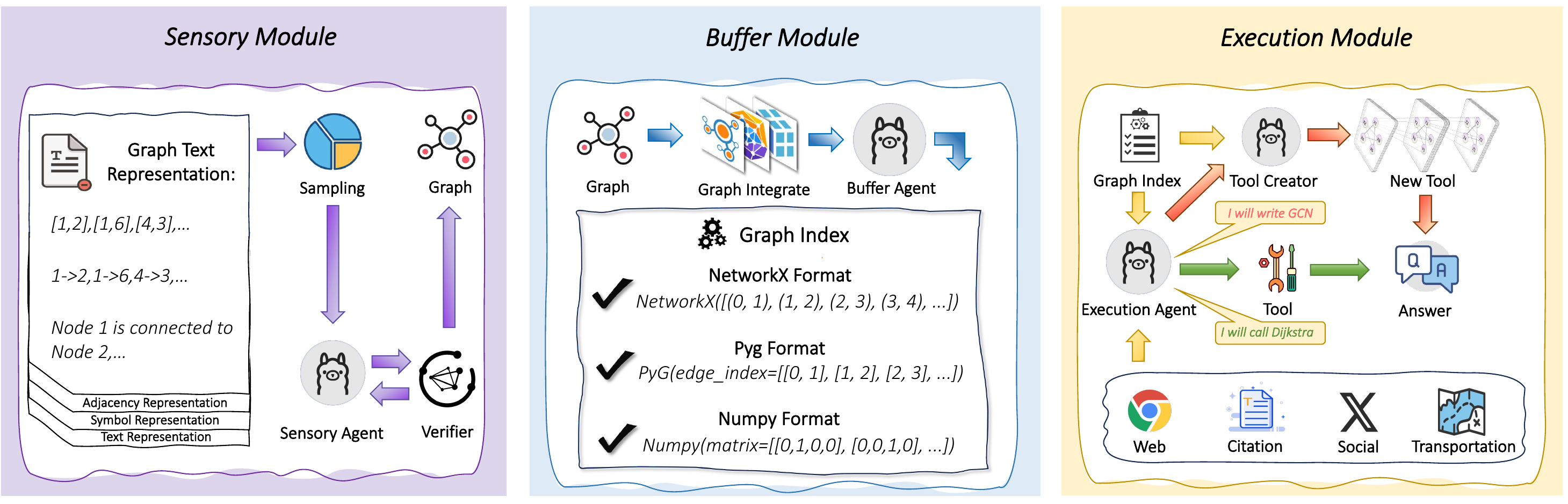}
  \caption{Overview of GraphCogent. Sensory Module (left) standardizes various graph text representations through subgraph sampling and conversion; Buffer Module (center) establishes cross-format data (e.g., NetworkX) integrating and indexing transformations; Execution Module (right) enables two reasoning modes: Execution Agent is employed for tool discrimination and implements tool calling for in-toolset tasks, Tool Creator handles out-toolset tasks based on tool creation.}
  \label{Chapter3_method}
\end{figure*}

 We propose GraphCogent, an agent-based framework inspired by the human working memory model to address the challenges of real-world graph reasoning tasks. Our framework consists of three specialized modules: Sensory Module (samples subgraphs and transform unstructured graph text representations into standardized adjacency list), Buffer Module (integrates graph data from Sensory Module and establishes data indices for diverse data formats), and Execution Module (combines tool calling and tool creation to address graph reasoning tasks). The overall pipeline of the GraphCogent is illustrated in Fig.~\ref{Chapter3_method}.

\subsection{Graph N-back Query}
Real-world graphs present two fundamental characteristics: (1) Large Graph Scale, benchmarks like NLGraph~\cite{DBLP:conf/nips/WangFHTHT23} use small graphs (less than 40 nodes), while real-world graph (e.g., Cora~\cite{DBLP:conf/aaai/CravenFMMNS98}) contains thousands of nodes and edges; (2) Varied Text Representation, different domains use distinct descriptions ("connected" in transportation or "followed" in social), and same text representation may appear in varied formats (adjacency lists, symbolic notations, linguistic descriptions).

We propose Graph N-back Query, a graph memory evaluation for LLM inspired by the human working memory assessment through N-back tasks~\cite{Kirchner_2006}: Using a 100-node graph (sampled from PeMS datasets~\cite{DBLP:journals/spm/ShumanNFOV13}) partitioned into 50-edge subsets \(E_t\), we test memory retention across three text representations by querying edge existence at turn \(t+N\). Experiments with Llama3.1-8B in Fig.~\ref{Chapter3_Nback} reveal significant forgetting across dialogues, demonstrating that LLMs’ inherent memory mechanisms fail to reliably handle large-scale graph reasoning. This highlights that LLMs' inherent context window limitations constrains their ability to maintain global graph information, rendering conventional step-by-step graph input approaches ineffective due to these working memory constraints.

\subsection{Sensory Module}
\label{Section 3.1}
The Sensory Module, which is proposed to address the dual challenges of graph scale and text representations, consists of two key components: Sensory Agent parses raw graph data via subgraph sampling and transform diverse text representations inputs into standardized adjacency lists; 
Graph Verifier serves as a supervisory mechanism for the Sensory Agent's transformation process, detecting both format deviations and misaligned conversions.

\textbf{Sensory Agent}: Experiments in Fig.~\ref{Chapter3_Representation} demonstrate that the granularity of sampling directly impacts LLMs' transformation performance. To achieve accurate text representation transformation, we define the Graph Efficiency Coefficient (GEC) to balance transformation accuracy (measured by Edit distance) against computational Cost (LLM token consumption):
\begin{equation}
    \text{GEC} = \text{Edit\_distance}(G, G') \times \frac{\sum_{i=1}^N (T_{\text{input}}^i + T_{\text{output}}^i)}{T_{\text{max}}},
    \label{eq:gec}
\end{equation}
where $\text{Edit\_distance}(G, G')$ measures the structural deviation between the original graph $G$ and the sampled graph $G'$; $\sum_{i=1}^N (T_{\text{input}}^i + T_{\text{output}}^i)$ quantifies the total token cost of LLM interactions, normalized by $T_{\text{max}}$ (maximum allowed tokens per LLM interaction). The algorithm for calculating Edit\_distance and the rationale for selecting token count as the Cost metric are detailed in Appendix~\ref{GEC}.

We prepare 100 graphs (300 edges each) to test four subgraph granularities (25-100 edges). Using Llama3.1-8B, we measure Edit distance and Cost to compute GEC. Results in Fig.~\ref{Chapter3_Representation} show that coarser granularity exceeds LLM's transformation capacity, increasing Edit distance, while finer granularity raises Cost through frequent LLM interactions. The GEC for graph reasoning tasks varies across models due to differences training or parameter scales, and for Llama3.1-8B, it achieves optimal GEC at approximately 50 edges.

\begin{figure*}
  \centering
  \includegraphics[width=\linewidth]{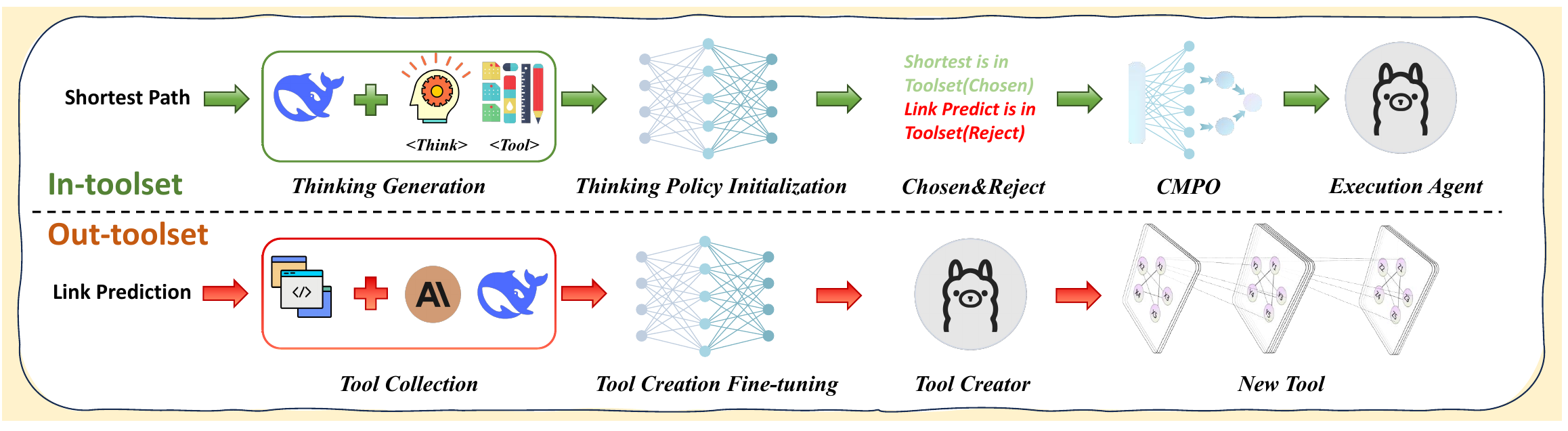}
  \caption{Overview of Execution Agent and Tool Creator training. In-toolset task (top) initializes a thinking policy from Think and Tool pairs, applies a CMPO method to refine Execution Agent’s tool capability discrimination; Out-toolset task (bottom) uses the fine-tuned Tool Creator to synthesize minimal task-specific tools, and backfills them into the common toolset.}
  \label{Main}
\end{figure*}

Guided by GEC, the Sensory Agent decomposes large graphs into optimally sized subgraphs to achieve accurate transformation. We employ heuristic prompts to elicit standardized adjacency lists regardless of input format (e.g., symbolic notations, linguistic descriptions). To handle both weighted and unweighted adjacency lists, we implement dual regular-expression templates that automatically match these formats; extracted edges are subsequently validated by the Graph Verifier. For text-represented graphs, we implement a sliding-window sampler whose window size is selected by GEC; by default with Llama3.1-8B we use 50 edges per window. Each window carries a fixed 5\% edge overlap to preserve cross-window structures at boundaries. For structured graph files (e.g., edgelist, txt, and json), we bypass partitioning and load them directly into the Buffer Module without windowing.

\textbf{Graph Verifier.} While GEC controls the granularity, LLM outputs can still introduce parsing noise during transformation. The Verifier checks quantity consistency (\(|V|\), \(|E|\)) and adjacency-list format compliance via regex parsing, and failed transformations trigger Sensory Agent retries. After per-window transformation, the Verifier merges subgraphs via a deterministic set-union with de-duplication and then runs a quick quantity-and-format check to ensure global consistency before dispatching any reasoning. By default, text graphs use a 50-edge window with 5\% overlap, and file-based graphs are not partitioned. Experiments on how the Graph Verifier enhances transformation reliability and ensures global consistency are provided in Appendix~\ref{Ablation_Sensory}.

\subsection{Buffer Module}
Current agent frameworks for graph reasoning lack storage mechanisms, relying instead on pre-processed files or delegating preprocessing to code generation. As Fig.~\ref{Chapter3_Representation} shows, even simple adjacency list transformations exhibit growing Edit distance with scale increasing, revealing LLMs’ declining reliability in graph comprehension. Moreover, graph paradigms diverge fundamentally in storage: algorithm tasks use OOP structures (e.g., NetworkX), link prediction requires tensor formats (e.g., PyG), and traffic prediction needs sparse matrices (e.g., NumPy). This difference overloads LLMs' working memory, thus resulting in inaccurate code generation, demanding both task-accurate implementations and cross-format transformation via reasoning.

Based on the human episodic buffer mechanism~\cite{DBLP:conf/aaai/GongWW24}, we construct the Buffer Module to serve as an intermediary that not only stores and integrates graph data but also establishes indices for different data formats. This module transforms raw graph text representations into multiple standardized formats: NetworkX for graph algorithms, NumPy for numerical operations, and PyG for tensor requirement tasks.

The Buffer Module begins by integrating adjacency lists from the Sensory Module to construct complete graph structures. To support diverse reasoning tasks, the raw graph data is transformed into multiple standardized data formats. For Structural Querying and Algorithmic Reasoning tasks, the data is preprocessed into NetworkX graph objects that preserve topological relationships and node attributes. For Predictive Modeling tasks, we implement data preprocessing including normalization and outlier treatment before encapsulating the graph as NumPy arrays or PyG tensors. To enable efficient access across these data format, we implement the Buffer Agent  to build data index that tracks key characteristics including data dimensionality, organizational schema, and metadata description. These indices are then passed to the Execution Module, allowing it to rapidly retrieve the most suitable graph format for each specific reasoning task.

\subsection{Execution Module}
Execution Module tackles code construction complexity and inefficiency through two collaborating components: Execution Agent and Tool Creator. The Execution Agent performs task analysis and tool invocation using a pre-built common toolset. It first assesses whether a problem instance is solvable within the toolset's capabilities via invoking existing tools; when covered, it directly executes the appropriate tool. For cases beyond current capabilities, it issues a creation request to the Tool Creator, which synthesizes a minimal task-specific tool and fills the pending invocation. By directly operating on the Buffer Module’s preprocessed data results, it avoids the fragility of full-code generation while retaining adaptability to complex requirements. Newly created tools can also be registered back into the toolset for future reuse.

\subsubsection{Execution Agent}
We adopt a two-stage instruction-tuning pipeline that couples (i) thinking policy initialization for robust tool invocation with (ii) capability-margin preference optimization (CMPO) for reliable tool capability discrimination. The first stage follows instruction-tuned policy learning with explicit thinking traces~\cite{DBLP:conf/iclr/HuSWALWWC22}; the second stage follows preference-based policy learning~\cite{DBLP:conf/nips/RafailovSMMEF23} to shape a likelihood-ratio margin around the tool capability boundary.

\noindent\textbf{Phase 1: Thinking Policy Initialization for Tool Invocation.}

We pair graph reasoning problems \textit{Prob} with toolset descriptions \textit{Toolset} to form composite queries \(\textit{Ques}=\langle\textit{Prob},\textit{Toolset}\rangle\). These are processed by DeepSeek-R1 to produce \(\textit{Ans}=\langle\textit{Think},\textit{Tool}\rangle\), where \textit{Think} exposes the reasoning trace to enhance routing stability and auditability, and \textit{Tool} indicates the selected tool. We curate pairs with correct invocation and optimize the maximum-likelihood objective:
\begin{equation}
\mathcal{L}_{\text{init}}=-\sum_{i=1}^{N}\log P_\theta\!\big(\langle\textit{Think},\textit{Tool}\rangle_i \,\big|\, \langle\textit{Prob},\textit{Toolset}\rangle_i\big),
\end{equation}
with the factorization:
\begin{equation}
\resizebox{0.9\linewidth}{!}{$
\log P_\theta(\textit{Ans}\mid\textit{Ques})
=\underbrace{\log P_\theta(\textit{Think}\mid\textit{Ques})}_{\text{thinking generation}}
+\underbrace{\log P_\theta(\textit{Tool}\mid\textit{Think},\textit{Ques})}_{\text{tool selection}}.
$}
\end{equation}

While the first stage provides strong invocation priors, Appendix~\ref{RQ3} shows a boundary effect: the agent may still over-rely on existing tools for tasks that actually require creation. This motivates a decision margin for the model’s tool capability. For later unification, we define a policy score relative to a frozen reference policy \(P_{\text{ref}}\) for subsequent preference optimization and tool-capability discrimination:
\begin{equation}
s_\theta(y\mid x)=\log P_\theta(y\mid x)-\log P_{\text{ref}}(y\mid x).
\end{equation}

\noindent\textbf{Phase 2: Capability-Margin Preference Optimization.}

To refine agent's tool capability discrimination, we adopt following two strategies.

\textit{(a) Action-segment scoring.}
\(\mathcal{M}_{\text{tool}}\) denotes the token indices belonging to the action segment (demarcated by special tool markers in data). We compute a policy score centered at a frozen reference policy \(P_{\text{ref}}\) but restricted to the action segment:
\begin{equation}
\resizebox{0.9\linewidth}{!}{$
\displaystyle
s_\theta^{\text{tool}}(y\mid x) 
= \sum_{t\in \mathcal{M}_{\text{tool}}}\!\log P_\theta(y_t\mid y_{<t},x)
- \sum_{t\in \mathcal{M}_{\text{tool}}}\!\log P_{\text{ref}}(y_t\mid y_{<t},x).
$}
\end{equation}
The score aggregates only the decision-critical tokens within the tool segment, isolating action selection from reasoning-style variance. We then quantify the pairwise preference between a preferred response \(R_w\) and a less-preferred response \(R_l\) by a likelihood-ratio margin:
\begin{equation}
\Delta s^{\text{tool}}
= s_\theta^{\text{tool}}(R_w\!\mid x) - s_\theta^{\text{tool}}(R_l\!\mid x) .
\end{equation}

\textit{(b) Capability-aware hard negatives.}
For each \(x\), we sample a candidate set \(\mathcal{C}_K(x)\) from the initialized policy and select a challenging negative via the reference action-segment likelihood:
\begin{equation}
R_l \;=\; \operatorname*{argmax}_{y\in \mathcal{C}_K(x)\setminus\{R_w\}} 
\ell_{\text{ref}}^{\text{tool}}(y\mid x),
\end{equation}
\begin{equation}
\ell_{\text{ref}}^{\text{tool}}(y\mid x)
\;=\; \sum_{t\in \mathcal{M}_{\text{tool}}} 
\log P_{\text{ref}}(y_t \mid y_{<t}, x).
\end{equation}
This selection aims to identify the most confusing incorrect answer for the agent based on the reference likelihood, forcing the agent to learn to distinguish fine-grained differences between them.

We then optimize a logistic objective with a fixed margin $m$:
\begin{equation}
\mathcal{L}_{\text{CMPO}}^{(\text{margin})}
= -\mathbb{E}_{(x,R_w,R_l)}
\Big[\log \sigma\big(\beta\,(\Delta s^{\text{tool}}-m)\big)\Big],
\end{equation}
which encourages the agent not only to prefer the correct tool over the negative, but to maintain a minimum confidence gap between them, thereby improving robustness near the decision boundary.

At execution time, we define a capability score \(\gamma(x)=\Delta s^{\text{tool}}\) and employ a two-threshold hysteresis rule to stabilize tool capability discrimination near the boundary, thereby further ensuring the stability and robustness of the agent’s decisions:
\begin{equation}
\text{route}(x)=
\begin{cases}
\mathcal{A}_{\text{invoke}}, & \gamma(x)\ge \tau_{\text{high}},\\[2pt]
\mathcal{A}_{\text{create}}, & \gamma(x)\le \tau_{\text{low}},\\[2pt]
\text{keep previous decision}, & \text{otherwise}.
\end{cases}
\end{equation}

\subsubsection{Tool Creator}

The Tool Creator handles missing-capability cases by generating only the tool body under a prompt template and returning it to fill the pending invocation. The template supplies the task description, admissible libraries and variable names, and the expected return object; outer interface and skeleton code is provided by the framework at runtime. By executing directly on the Buffer Module’s preprocessed data structures, this design keeps generation reliable and avoids brittle full-file synthesis.

We curate a corpus from classical graph algorithms (e.g., PageRank) and neural models (e.g., GCN, GAT) on GitHub. Using LLMs (DeepSeek-v3~\cite{DBLP:journals/corr/abs-2501-12948}, GPT-4o~\cite{DBLP:journals/corr/abs-2303-08774}, Claude-3.7~\cite{TheC3}), we synthesize code and retain only samples that execute successfully. For each retained sample, we extract the body segment between special body marker. On the input side, we concatenate the original query with its task template to form $\textit{Ques}^{\text{Tpl}}=\langle\textit{Ques},\mathcal{T}\rangle$, and keep a minimal contextual scaffold $c_i$ (e.g., necessary imports, a function signature, lightweight placeholders) to induce body-only completion.

\textbf{Tool Creation Fine-tuning.} We optimize a body-only maximum likelihood objective:
\begin{equation}
\label{eq:training_objective}
\mathcal{L}_{\text{train}}
= -\sum_{i=1}^{N}\ \sum_{t=1}^{M_{\text{body},i}}
\log P_\theta\!\big(y_{i,t}\,\big|\, y_{i,<t},\ \textit{Ques}^{\text{Tpl}}_i,\ c_i\big),
\end{equation}
where $N$ is the number of validated examples, $M_{\text{body},i}$ denotes the number of tokens in the body segment of example $i$ (we reindex body tokens as $t=1,\dots,M_{\text{body},i}$), $y_{i,t}$ is the $t$-th target token within the body. This objective focuses learning on the implementation logic while leaving interface details to the framework. At inference, the template provides the necessary task and contextual cues; the Tool Creator emits the body segment to satisfy the Execution Agent’s pending invocation, which the framework assembles with the runtime skeleton and executes to produce final results.

\subsubsection{Training Settings}
We use Llama3.1-8B as the backbone for all agents. For capability-aware negatives, the candidate set size is $K=8$. Two-thresholds are set to $\tau_{\text{low}} = 0.55$ and $\tau_{\text{low}} = 0.45$. Both Execution Agent and Tool Creator are trained on Graph4real dataset. The training set employs 4 of 20 prompt templates, with their corresponding tasks excluded from the test set to ensure fair evaluation. More detailed settings can be founed in Appendix~\ref{Experimental Settings}.

\section{Experiments}
\subsection{Experimental Settings}
\label{secExperimental Settings}

\textbf{Dataset.} We evaluate on Graph4real, a real-world benchmark with graphs from 40 to 1,000 nodes spanning transportation, social, web, and citation domains. The dataset includes both in-toolset and out-toolset tasks, with 500 test instances per task and scale (across domains). To ensure fair comparisons, we use standard adjacency lists for graph text representation for main experiments in Table~\ref{Table1}. Additionally, we introduce public datasets: NLGraph~\cite{DBLP:conf/nips/WangFHTHT23}, Talk like a Graph~\cite{DBLP:journals/corr/abs-2310-04560}, GraphWiz~\cite{DBLP:journals/corr/abs-2402-16029}, GraphArena~\cite{DBLP:conf/iclr/TangZLCL25}, GraCoRe~\cite{DBLP:conf/coling/YuanLWQ25} and GraphInstruct~\cite{DBLP:journals/corr/abs-2403-04483}.

\noindent\textbf{Baseline Methods.} We compare three types of methods: (1) \textit{Text-based methods} including GPT-4o~\cite{DBLP:journals/corr/abs-2303-08774}, Claude-3.7~\cite{TheC3}, and DeepSeek-R1~\cite{DBLP:journals/corr/abs-2501-12948} with 2-shot CoT prompting. (2) \textit{Tool-based methods} including Graph-Toolformer~\cite{DBLP:journals/corr/abs-2304-11116}, GraphTool-Instruction~\cite{DBLP:conf/kdd/Wang0CZQ25} on Llama3.1-8B~\cite{DBLP:journals/corr/abs-2302-13971} and GPT-4o with Function Calling~\cite{DBLP:journals/corr/abs-2303-08774}; (3) \textit{Agent-based methods} are the state-of-the-art GraphTeam~\cite{DBLP:journals/corr/abs-2410-18032} framework with GPT-4o-mini~\cite{DBLP:journals/corr/abs-2303-08774} and reproduced GCoder~\cite{zhang2024gcoder} with Llama3.1-8b~\cite{DBLP:journals/corr/abs-2302-13971}. All LLMs' versions and settings are in Appendix~\ref{Experimental Settings}.

\noindent\textbf{Evaluation Metrics.} We use accuracy for all tasks except traffic flow prediction (measured by MAE).

\begin{table*}[t]
\centering
\small
\caption{Performances of GraphCogent and other baselines on Graph4real in-toolset tasks. \textbf{Note:} Graph-TF, GraphTool-Ins ,and GPT4o-FC are abbreviations for Graph-Toolformer, GraphTool-Instruction,and GPT4o-Function Calling.}
\label{Table1}
\begin{tabular}{@{}l|l|ccccccccc|c@{}}
\toprule
\multicolumn{1}{@{}c|}{\textbf{Method}} &
\multicolumn{1}{c|}{\textbf{Model}} &
\makecell{Edge\\Existence} &
\makecell{Edge\\Count} & \makecell{Triangle\\Count} & \makecell{Shortest\\Path} & \makecell{Path\\Existence} & \makecell{Cycle\\Detection} & \makecell{Node\\Existence} & \makecell{Node\\Count} & \makecell{Degree\\Count} & \makecell{Average} \\
\midrule
\rowcolor{gray!15} \multicolumn{12}{@{}l}{\textbf{Small-Scale (40 Nodes)}} \\
\multirow{3}{*}{Text} & GPT-4o~\cite{DBLP:journals/corr/abs-2303-08774}          & \textbf{100} & 22.2 & 20.6 & 25.8 & 77.2 & 78.2 & \textbf{100} & 84.2 & 29.8 & 59.8 \\ 
& Claude-3.7~\cite{TheC3}      & 99.2 & 22.4 & 21.0 & 21.2 & 79.8 & 81.8 & \textbf{100} & 80.2 & 32.2 & 59.8 \\ 
& DeepSeek-R1~\cite{DBLP:journals/corr/abs-2501-12948}     & \textbf{100} & 32.6 & 26.6 & 30.8 & 86.6 & 86.8 & \textbf{100} & \underline{90.4} & 31.0 & 65.0 \\ 
\cmidrule{1-12}

\multirow{3}{*}{Tool} & Graph-TF~\cite{DBLP:journals/corr/abs-2304-11116}       & 48.0 & 75.6 & 71.0 & 63.8 & 44.2 & 66.8 & 50.2 & 77.2 & 50.0 & 60.8 \\ 
& GraphTool-Ins~\cite{DBLP:conf/kdd/Wang0CZQ25}   & 94.2 & 98.0 & 97.8 & 98.2 & 96.8 & \underline{99.0} & \textbf{100} & \textbf{100} & 97.8 & 98.0 \\ 
& GPT4o-FC~\cite{DBLP:journals/corr/abs-2303-08774}         & 95.0 & \textbf{99.2} & \underline{98.4} & \textbf{99.0} & \textbf{99.2} & \textbf{100} & \textbf{100} & \textbf{100} & \underline{99.0} & \underline{98.9} \\ 
\cmidrule{1-12}

\multirow{3}{*}{Agent} & GraphTeam~\cite{DBLP:journals/corr/abs-2410-18032}      & 86.6 & 84.8 & 82.8 & 80.2 & 97.6 & 80.0 & 96.8 & 70.8 & 97.6 & 86.4 \\
& GCoder~\cite{zhang2024gcoder} & 82.0 & 84.6 & 83.4 & 82.8 & 90.6 & 80.6 & 88.0 & 74.0 & 87.8 & 83.8 \\
& \textbf{GraphCogent} & \underline{99.4} & \underline{98.2} & \textbf{99.0} & \underline{98.6} & \underline{97.8} & \textbf{100} & \underline{99.2} & \textbf{100} & \textbf{100} & \textbf{99.1} \\
\midrule

\rowcolor{gray!15} \multicolumn{12}{@{}l}{\textbf{Middle-Scale (100 Nodes)}} \\
\multirow{3}{*}{Text} & GPT-4o~\cite{DBLP:journals/corr/abs-2303-08774}          & 90.4 & 6.8 & 5.4 & 7.2 & 54.0 & 67.6 & 99.8 & 58.6 & 6.8 & 44.1 \\ 
& Claude-3.7~\cite{TheC3}      & 93.0 & 6.2 & 4.8 & 6.8 & 55.8 & 73.0 & 99.2 & 52.2 & 7.2 & 44.2 \\ 
& DeepSeek-R1~\cite{DBLP:journals/corr/abs-2501-12948}     & \underline{98.0} & 7.4 & 6.2 & 8.0 & 60.6 & 76.6 & \textbf{100} & 60.4 & 7.4 & 47.2 \\ 
\cmidrule{1-12}
\multirow{3}{*}{Tool} & Graph-TF~\cite{DBLP:journals/corr/abs-2304-11116}       & 43.2 & 68.0 & 63.8 & 57.4 & 39.8 & 60.0 & 45.2 & 69.4 & 45.0 & 54.6 \\ 
& GraphTool-Ins~\cite{DBLP:conf/kdd/Wang0CZQ25}   & 84.8 & 89.0 & 88.0 & 88.4 & 89.0 & 89.0 & 90.6 & 92.0 & 88.0 & 88.8 \\ 
& GPT4o-FC~\cite{DBLP:journals/corr/abs-2303-08774}         & 85.4 & \underline{89.2} & \underline{88.6} & \underline{89.0} & \underline{89.2} & \underline{89.4} & 97.0 & \underline{94.2} & 89.0 & \underline{90.1} \\ 
\cmidrule{1-12}
\multirow{3}{*}{Agent} & GraphTeam~\cite{DBLP:journals/corr/abs-2410-18032}      & 91.6 & 80.2 & 73.0 & 65.8 & 93.2 & 69.6 & 89.6 & 65.4 & \underline{95.2} & 80.4 \\ 
& GCoder~\cite{zhang2024gcoder} & 80.0 & 68.0 & 60.8 & 68.2 & 82.4 & 71.4 & 78.2 & 67.4 & 82.8 & 73.2 \\
& \textbf{GraphCogent} & \textbf{98.2} & \textbf{97.8} & \textbf{98.8} & \textbf{98.2} & \textbf{97.4} & \textbf{99.0} & \underline{99.4} & \textbf{97.6} & \textbf{99.8} & \textbf{98.5} \\ 
\midrule

\rowcolor{gray!15} \multicolumn{12}{@{}l}{\textbf{Large-Scale (1000 Nodes)}} \\
Agent & \textbf{GraphCogent} & \textbf{97.0} & \textbf{97.4} & \textbf{98.6} & \textbf{98.0} & \textbf{97.2} & \textbf{97.0} & \textbf{96.8} & \textbf{97.2} & \textbf{99.6} & \textbf{97.6} \\
\bottomrule
\end{tabular}
\end{table*}

\begin{table*}[htbp]
\centering
\small
\caption{Performance evaluation on out-toolset tasks. \textbf{Note:} Except for the traffic prediction (which values represent MAE), all other values represent Accuracy.}
\label{Table2}
\begin{tabular}{@{}l|ccc|ccc|ccc|ccc@{}}
\toprule
& \multicolumn{3}{c|}{\textbf{Web}} & \multicolumn{3}{c|}{\textbf{Transportation}} & \multicolumn{3}{c|}{\textbf{Social}} & \multicolumn{3}{c}{\textbf{Citation}} \\ 
\cmidrule(lr){2-4} \cmidrule(lr){5-7} \cmidrule(lr){8-10} \cmidrule(lr){11-13}
\multicolumn{1}{c|}{\textbf{Method}} 
& Common & PageRank & Link 
& Max & Diameter & \textit{Traffic} 
& Maxcore & Connect. & Link 
& Reference & Cluster. & Node \\ 
& Neighbor & Calcu. & Pred. 
& Flow & Calcu. & \textit{Pred.} 
& Calcu. & Compo. & Pred. 
& Match & Coeff. & Class. \\ 
\midrule
GraphTeam & 51.2 & 76.8 & 71.6 & 72.4 & 61.6 & \textit{96.7} & 69.8 & 72.0 & 76.4 & 78.8 & 76.4 & 72.0 \\
GCoder & 49.0 & 74.0 & 68.5 & 70.0 & 58.8 & \textit{99.2} & 67.2 & 69.0 & 72.0 & 75.0 & 74.0 & 68.0 \\
GraphCogent & 88.8 & 92.6 & 86.4 & 89.6 & 92.4 & \textit{35.1} & 91.6 & 87.6 & 85.2 & 90.0 & 85.2 & 89.6 \\
\bottomrule
\end{tabular}
\end{table*}

\subsection{Main Result}
\label{Main Result}

The performance results are reported in Table~\ref{Table1}. Our method significantly outperforms all baseline methods across toolset-covered tasks. As for text-based methods, they demonstrate advantages on small-scale simple Structural Querying tasks (e.g., DeepSeek-R1 achieves 100\% on edge existence). However, these methods show significant performance degradation when dealing with problems requiring multi-step reasoning (e.g., only 30.8\% accuracy on shortest path), and this issue becomes more pronounced as the scale increases. Tool-based approaches such as GPT4o-FC demonstrates robust performance across all tasks on small-scale graphs by leveraging GPT4o's powerful reasoning capabilities. However, they suffer 10\% performance degradation when scaling from 40 to 100 nodes. The agent-based GraphTeam and GCoder demonstrate limitations in code generation across multiple tasks. The complex data preprocessing and reasoning requirements overload the model's working memory, resulting in lower performance. In contrast, our method maintains state-of-the-art performance across all scales with less than 1\% variance. Notably, while no baseline could handle large-scale graphs, we have only included the results of our approach.

\begin{table*}[h]
\small
\centering
\caption{Reasoning token comparison between agent baselines. \textbf{Note:} Input and Output are abbreviation for average input tokens and average output tokens respectively.}
\label{tab:comparison}
\begin{tabular}{@{}l|ccc|ccc|ccc|ccc@{}}
\toprule
 & \multicolumn{6}{c|}{\textbf{In-Toolset Tasks}} & \multicolumn{6}{c}{\textbf{Out-Toolset Tasks}} \\
 \cmidrule(lr){2-4} \cmidrule(lr){5-7} \cmidrule(lr){8-10}  \cmidrule(lr){11-13}
 \multicolumn{1}{c|}{\textbf{Method}} & \multicolumn{3}{c|}{Shortest Path} & \multicolumn{3}{c|}{Cycle Detection} & 
 \multicolumn{3}{c|}{Max Flow} & \multicolumn{3}{c}{Link Prediction} \\
 \cmidrule(lr){2-4} \cmidrule(lr){5-7} \cmidrule(lr){8-10} \cmidrule(lr){11-13}
 & Input & Output & Accuracy & Input & Output & Accuracy & Input & Output & Accuracy & Input & Output & Accuracy \\
\midrule
GraphTeam & 4248 & 3285 & 65.8 & 4207 & 2901 & 69.6 & 4122 & 3023 & 72.4 & 4423 & 3820 & 76.4 \\
GCoder & 3782 & 1051 & 68.2 & 3640 & 620 & 71.4 & 3927 & 1181 & 70.0 & 3855 & 1909 & 72.0 \\
GraphCogent & 2794 & 615 & 98.2 & 2533 & 595 & 99.0 & 3009 & 1044 & 89.6 & 3128 & 1244 & 90.2 \\
\bottomrule
\end{tabular}
\end{table*}

\begin{table*}[h]
\small
\centering
\caption{Reasoning time comparison between GraphCogent and other baselines.}
\label{tab:timebenchmark}
\begin{tabular}{l|cc|cc|cc|cc|cc}
\toprule
\multicolumn{1}{c|}{\multirow{2}{*}{\textbf{Methods}}} & \multicolumn{2}{c|}{Shortest Path} & \multicolumn{2}{c|}{Cycle Detection} & \multicolumn{2}{c|}{Triangle Count} & \multicolumn{2}{c|}{Path Existence} & \multicolumn{2}{c}{Link Prediction} \\
\cmidrule(lr){2-3} \cmidrule(lr){4-5} \cmidrule(lr){6-7} \cmidrule(lr){8-9} \cmidrule(lr){10-11}
 & Time & Accuracy & Time & Accuracy & Time & Accuracy & Time & Accuracy & Time & Accuracy \\
\midrule
GPT4o-Function Calling & 12.2s & 89.0 & 10.7s & 89.4 & 11.3s & 88.6 & 12.9s & 89.2 & \slash & \slash \\
Graph-Toolformer & 18.1s & 57.4 & 16.6s & 60.0 & 18.9s & 63.8 & 17.3s & 39.8 & \slash & \slash \\
GraphTool-Instruction & 19.8s & 88.4 & 18.3s & 89.0 & 20.1s & 88.0 & 19.2s & 89.0 & \slash & \slash \\
GraphTeam & 58.1s & 65.8 & 56.6s & 69.6 & 57.3s & 73.0 & 58.9s & 93.2 & 59.5s & 76.4 \\
GCoder & 21.5s & 68.2 & 17.5s & 71.4 & 17.4s & 60.8 & 13.3s & 82.4 & 27.7s & 72.0 \\
GraphCogent & 22.8s & 98.2 & 23.3s & 99.0 & 20.1s & 98.8 & 19.2s & 97.4 & 24.0s & 85.2 \\
\bottomrule
\end{tabular}
\end{table*}

We then evaluate GraphCogent's performance on out-toolset tasks against GraphTeam and GCoder in Table~\ref{Table2} on middle-scale graphs. Focusing on agent capabilities for handling uncovered tasks, we omit tool-based methods from comparison as they inherently lack these functionalities. Our method demonstrates superior adaptability, achieving 20\% improvements over GraphTeam and GCoder across four domains.

\begin{figure}
    \centering
\includegraphics[width=\linewidth]{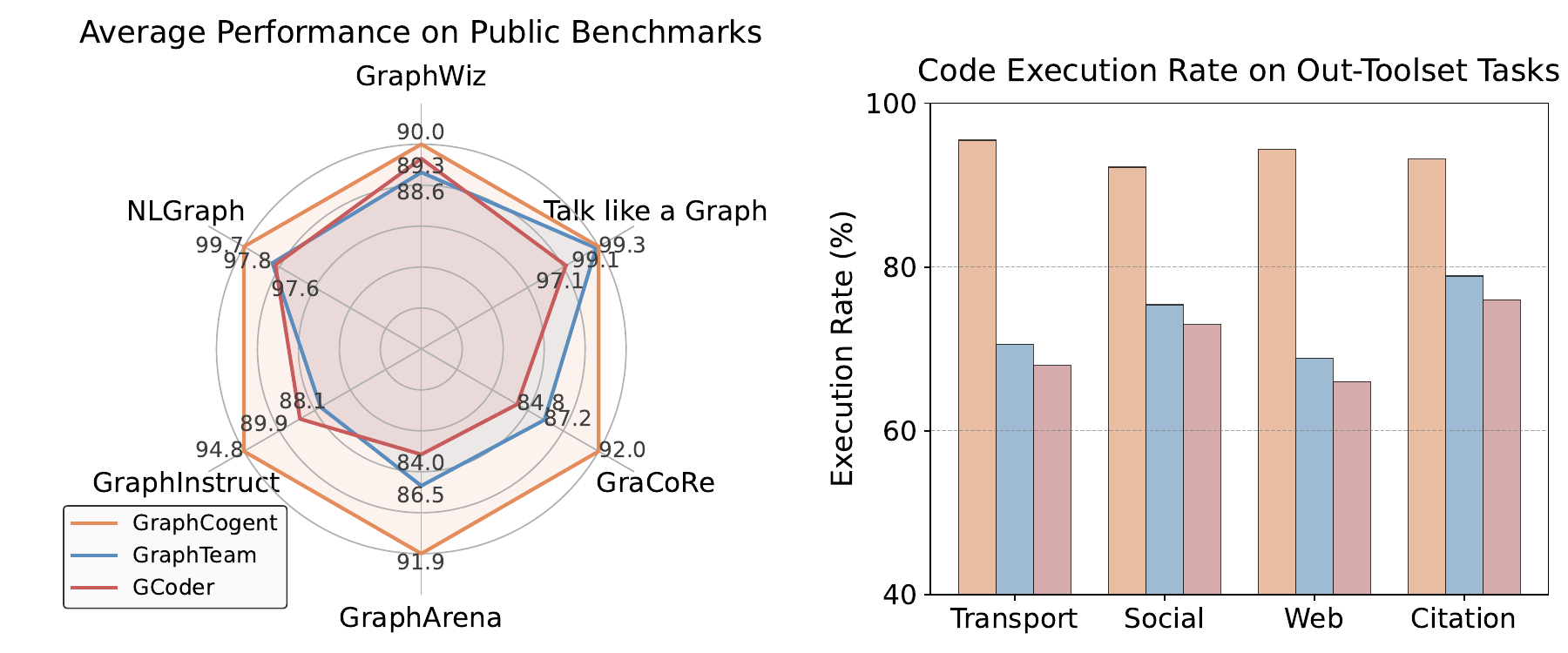}
\caption{
    Average Performance on Public Benchmarks (right). Code execution rate on out-toolset tasks (left).
}
\label{Chapter4_Code Execution Rate Statistics}
\end{figure}

We further conduct a systematic comparison among agent-based approaches on public benchmarks (right part of Fig.~\ref{Chapter4_Code Execution Rate Statistics}). For fair comparison, the evaluation is conducted on the intersection of tasks jointly covered by three agent-based methods. The results demonstrate that GraphCogent maintains strong performance across all six public datasets. We attribute this to our working memory model inspired pipeline, which leads to lower execution-error rates and more efficient runs. This is further supported by the Code Execution Rate statistics (left part of Fig.~\ref{Chapter4_Code Execution Rate Statistics}), where our method achieves higher one-pass success rates than GraphTeam’s three-retry mechanism and GCoder’s compiler-feedback–based reinforcement tuning. These findings are consistent with our analysis of working memory constraints: when an LLM must simultaneously retain complex graph topology, perform multi-step reasoning, and synthesize executable code, it tends to degrade in both accuracy and executability.

Experiments on effectiveness evaluation of our approach demonstrate the superiority in two key aspects. First, our token consumption analysis in Table~\ref{tab:comparison} shows consistent improvements across all tasks, with significantly reduced input token consumption and substantial output token savings over GraphTeam (approximately 80\% for in-toolset tasks and 30\% for out-toolset tasks) while maintaining high accuracy. Second, we compare Tool-based and Agent-based methods on reasoning time in Table~\ref{tab:timebenchmark}. While GPT4o-FC's API implementation demonstrates significant time advantages over local open-source LLMs, our local method GraphCogent achieves competitive speed while maintaining superior accuracy. The additional reasoning time in our framework mainly stems from multi-round sampling in the Sensory module. In contrast, although GraphTeam (based on GPT4o-mini) benefits from faster API inference, its performance is compromised by working memory overload in complex tasks and the overhead of its three-retry mechanism, making its direct code generation approach both slower and less accurate.


\begin{table}
    \caption{Accuracy on larger graph.}
    \centering
    \small
    \adjustbox{max width=\linewidth}{
    \begin{tabular}{l|cccccc}
    \toprule
    \multirow{2}{*}{\textbf{Scale}} & Shortest & Cycle & Triangle & Path & PageRank & Max \\
     & Path & Detect. & Count & Exist. & Calcu. & Flow \\
    \midrule
    2000  & 97.0 & 97.0 & 98.6 & 97.0 & 93.0 & 89.6 \\
    5000  & 98.2 & 97.0 & 98.8 & 98.0 & 90.8 & 87.8 \\
    10000 & 97.4 & 96.8 & 99.0 & 97.0 & 89.8 & 88.4 \\
    \bottomrule
    \end{tabular}
    }
    \label{tab:accuracy}
\end{table}

\begin{table}[h]
\small
\centering
\caption{Framework adaptability across different LLMs.}
\label{tab:adaptability}
\begin{tabular}{l|cccc}
\toprule
\multirow{2}{*}{\textbf{Base Model}} & Shortest & Cycle & Max & Link \\
 & Path & Detect. & Flow & Pred. \\
\midrule
Naive GLM4-9B & 0.0 & 68.4 & 0.0 & 48.6 \\
GraphCogent (GLM) & 96.2 & 96.8 & 81.4 & 82.2 \\
\midrule
Naive Qwen2.5-7B & 0.0 & 56.4 & 0.0 & 51.0 \\
GraphCogent (Qwen) & 92.2 & 90.4 & 91.2 & 87.2 \\
\midrule
Naive GPT-4o & 7.2 & 67.6 & 10.2 & 62.4 \\
GraphCogent (GPT-4o) & 100 & 100 & 99.2 & 90.8 \\
\midrule
Naive DeepSeek-R1 & 8.0 & 76.7 & 9.6 & 60.4 \\
GraphCogent (DeepSeek-R1) & 100 & 100 & 99.2 & 91.2 \\
\bottomrule
\end{tabular}
\end{table}

Although our experiments primarily focus on the Llama3.1-8B, the framework's modular design ensures compatibility with both open-source and closed-source LLMs. Table~\ref{tab:adaptability} reveals our framework's adaptability. For open-source models like GLM4-9B-chat and Qwen2.5-7B-coder-Instruct, our framework significantly enhances their reasoning capabilities especially for complex tasks requiring multi-step graph reasoning, improving performance from near-zero accuracy in tasks such as Shortest Path and Max Flow to high accuracy. Second, based on the framework's instruction without training, massive-scale LLMs like GPT-4o and DeepSeek-R1 achieve high performance with in-toolset tasks and out-toolset, demonstrating the universal applicability of our framework design.

The scalability evaluation in Table~\ref{tab:accuracy} on larger graphs further demonstrates the effectiveness of our framework, where it maintains stable performance across various tasks. 
Through ablation studies in Appendix~\ref{Ablation Study}, we also examine the contributions of each component.

\section{Conclusion}
In this work, we proposed GraphCogent, a novel agent framework that enhances LLMs' graph reasoning capabilities by mitigating their working memory constraints. Inspired by human cognitive architecture, GraphCogent integrates three key modules: Sensory Module (standardizing diverse graph text representations), Buffer Module (integrating and indexing graph data), and Execution Module (combining tool-calling and tool-creation reasoning). To evaluate LLMs on real-world graph reasoning tasks, we introduce Graph4real, a benchmark featuring large-scale graphs for four real-world domains. Experiments show that GraphCogent achieves 20\% higher than existing state-of-the-art agent-based method while reducing over 30\% token usage.


\bibliographystyle{ACM-Reference-Format}
\bibliography{sample-base}

\appendix

\section{Additional Experimental Settings}
\label{Experimental Settings}

\textbf{Corresponding LLM versions.} All corresponding versions of the LLMs are presented in Table~\ref{LLM Version}.

\begin{table}[h]
\scriptsize
\centering
\caption{Baseline methods and corresponding models.}
\begin{tabular}{c|cc}
\toprule
LLM Type & Method & Base Model \\ \midrule
Text-based & Two-shot prompt~\cite{DBLP:conf/nips/Wei0SBIXCLZ22} & GPT-4o-2024-08-06~\cite{DBLP:journals/corr/abs-2303-08774} \\
& Two-shot prompt~\cite{DBLP:conf/nips/Wei0SBIXCLZ22} & Claude-3.7-sonnet-20250219~\cite{TheC3} \\
& Two-shot prompt~\cite{DBLP:conf/nips/Wei0SBIXCLZ22} & DeepSeek-R1~\cite{DBLP:journals/corr/abs-2501-12948} \\ \midrule
Tool-based & Graph-Toolformer~\cite{DBLP:journals/corr/abs-2304-11116} & Llama3.1-8B-Instruct~\cite{DBLP:journals/corr/abs-2302-13971} \\
& GraphTool-Instruction~\cite{DBLP:conf/kdd/Wang0CZQ25} & Llama3.1-8B-Instruct~\cite{DBLP:journals/corr/abs-2302-13971} \\
& Function Calling~\cite{DBLP:journals/corr/abs-2303-08774} & GPT-4o-2024-08-06~\cite{DBLP:journals/corr/abs-2303-08774} \\ \midrule
Agent-based & GraphTeam~\cite{DBLP:journals/corr/abs-2410-18032} & GPT-4o-mini-2024-07-18~\cite{DBLP:journals/corr/abs-2303-08774} \\
& GCoder~\cite{zhang2024gcoder} & Llama3.1-8B-Instruct~\cite{DBLP:journals/corr/abs-2302-13971} \\
& GraphCogent & Llama3.1-8B-Instruct~\cite{DBLP:journals/corr/abs-2302-13971} \\ \bottomrule
\end{tabular}
\label{LLM Version}
\end{table}

\textbf{Setting.} All open-source LLM methods were trained on 8$\times$NVIDIA A800 (80G) GPUs and evaluated on 8$\times$NVIDIA 4090 (24G) GPUs. Default parameters were used with temperature=0.7 and top\_p=1. Instruction fine-tuning and CMPO training were conducted with the Llama-Factory framework. For instruction fine-tuning, we used a learning rate of 1e-5, warmup ratio 0.1, batch size 4, cosine scheduler, and 5 epochs. For CMPO, we used a learning rate of 5e-6, batch size 1 with 4 gradient accumulation steps, warmup ratio 0.1, cosine scheduler, and 5 epochs. Closed-source LLMs (OpenAI, Anthropic) were accessed via official APIs with default settings (temperature=0.7, top\_p=1).

\section{Ablation Study}
\label{Ablation Study}

We conduct ablation studies to evaluate each module and its key components through the following research questions:




\subsection{Ablation Studies on Sensory Module}
\label{Ablation_Sensory}

\textbf{RQ1: How does the Graph Verifier component enhance transformation reliability and ensure global consistency?}

As shown in Fig.~\ref{Chapter3_Representation} even based on GEC optimization, the inherent uncertainty in LLM outputs can still introduce parsing inaccuracies during graph transformation. Table \ref{tab:verifier_impact} further demonstrates the critical role of the Graph Verifier in ensuring reliable graph transformation and global consistency. 

We randomly sample 100 graphs (each containing 100 nodes with 300 edges) from PeMS dataset, partitioning each graph into 50-edges subsets with different graph text representations (i.e., adjacency list, symbolic notation, and linguistic description). We evaluate edit distance during transformation and assess post-merge global consistency using two criteria: (i) Global Edit Distance (GED), the edit distance between the original graph \(G\) and the merged graph \(\hat{G}\); and (ii) Boundary\textendash Edge Recall (BER), the recall of edges whose endpoints lie in different windows or touch a window boundary. The results show that introducing the Graph Verifier reduces Edit distance by approximately 90\%, while maintaining operational efficiency with fewer than 10 verification triggers per transformation. The results show that introducing the Graph Verifier reduces structural errors by approximately 90\%, while maintaining operational efficiency with fewer than 10 verification triggers. This retry frequency (occurring in approximately 600 LLM calls) remains acceptable while significantly reducing edit distance.

\begin{table}[h]
\small
\centering
\caption{Impact of Graph Verifier on transformation reliability and global consistency. Edit Distance is computed by summing the edit distances of all windows within each graph.}
\label{tab:verifier_impact}
\resizebox{\linewidth}{!}{
\begin{tabular}{lccccc}
\toprule
Representation Type & Edit Dist. (Before) & Edit Dist. (After) & GED & BER (\%) & Verifier Triggers \\
\midrule
Adjacency List         & $44.8 \pm 7.6$ & $5.2 \pm 1.1$ & $0.8 \pm 0.2$ & $99.8 \pm 0.06$ & $6.2 \pm 1.0$ \\
Symbolic Notation      & $57.3 \pm 8.9$ & $5.9 \pm 1.2$ & $1.1 \pm 0.4$ & $99.7 \pm 0.08$ & $9.0 \pm 1.3$ \\
Linguistic Description & $65.9 \pm 9.1$ & $5.1 \pm 1.0$ & $1.0 \pm 0.3$ & $99.8 \pm 0.07$ & $9.1 \pm 1.2$ \\
\bottomrule
\end{tabular}
}
\end{table}

\vspace{-1em}
\subsection{Ablation Studies on Buffer Module}

\textbf{RQ2: How does the Buffer Mechanism enhance the framework's reasoning capability?} 

We select four tasks within Graph4real on middle-scale. The experimental results in Table~\ref{tab:ablation_buffer} show that removing the Buffer Module causes substantial performance decreases across all tasks. This result evidences the Buffer Module's critical role in addressing LLMs' working memory constraints for graph reasoning. Without the Buffer Module, when the graph scale exceeds the capacity of LLMs' working memory, the LLMs exhibit substantial information loss regarding graph topology.

\begin{table}[h]
\scriptsize
\centering
\caption{Performance impact of Buffer Module removal.}
\label{tab:ablation_buffer}
\resizebox{\linewidth}{!}{
\begin{tabular}{l*{4}{c}}
\toprule
\multicolumn{1}{c}{\textbf{Variants}} & Shortest Path & Cycle Detection & Max Flow & Link Prediction \\
\midrule
GraphCogent & 98.2 & 99.0 & 89.6 & 85.2 \\
GraphCogent w/o Buffer & 68.4 & 69.2 & 53.8 & 54.2 \\
\bottomrule
\end{tabular}
}
\end{table}

\subsection{Ablation Studies on Execution Module}
\label{RQ3}

\textbf{RQ3: Does Thinking Policy Initialization (TPI) improve Execution Agent's tool selection?} 

We select four tasks within Graph4real on middle-scale. Three versions of the Execution Agent are evaluated: the original untuned agent (Llama3.1-8B), agent fine-tuned using the Naive SFT method, and agent fine-tuned with our Thinking Policy Initialization (TPI). The Naive SFT approach involves training Llama3.1-8B by directly mapping problem statements to their outputs, where each output consists of a brief analytical statement containing the tool selection result. As shown in Table~\ref{tab:ablation_sft}, the TPI method shows consistent performance advantages across different tasks. For in-toolset tasks, TPI achieves a 5\% performance improvement over the naive approach, as the explicit reasoning chains enable the agent to better understand task intent and select appropriate tools. Through analysis of the reasoning process, we observe that compared to Naive SFT's tendency to directly output corresponding tools (i.e., simply fitting tool results), the TPI teaches the agent to comprehend task intentions and demonstrates reasonable reasoning paths, leading to correct tool selection. This advantage becomes even more significant for out-toolset tasks, where the TPI shows substantially better generalization capabilities than naive method.

\vspace{-1em}
\begin{table}[h]
\scriptsize
\centering
\caption{Impact of Thinking Policy Initialization on Execution Agent performance.}
\label{tab:ablation_sft}
\resizebox{\linewidth}{!}{
\begin{tabular}{l*{4}{c}}
\toprule
\multicolumn{1}{c}{\textbf{Training Method}} & Shortest Path & Cycle Detection & Max Flow & Link Prediction \\
\midrule
Llama3.1-8B (Base) & 78.4 & 79.2 & 35.8 & 36.2 \\
Llama3.1-8B w/ SFT & 89.3 & 90.2 & 43.8 & 44.2 \\
Llama3.1-8B w/ TPI & 98.2 & 99.0 & 70.2 & 65.0 \\
\bottomrule
\end{tabular}
}
\end{table}

However, the TPI still has limitations. Through analyzing the reasoning results for out-toolset tasks like Max Flow and Link Prediction, we identify a tendency of the LLM to forcibly associate tools. For instance, although no single tool can solve the Max Flow problem, unlike the Naive SFT-tuned LLM, which randomly generates a tool or calls a non-existent tool, the TPI-tuned LLM attempts to combine tools to derive a solution. This preference actually contradicts practical decision-making. In this task, we would prefer the agent to directly generate the task-specific tool of Max Flow through Tool Creator rather than repeatedly interact with tools of Execution Agent. This observation motivates our implementation of CMPO alignment to refine the Execution Agent's preferences.

\textbf{RQ4: Why employ Capability-Margin Preference Optimization (CMPO) for tool capability discrimination?} 

Building upon the findings from~\ref{RQ3}, the core objective of CMPO alignment is to refine the Execution Agent's tool selection preferences for improving toolset discrimination capabilities. Although tasks are not strictly binary (i.e., solvable either by tools or not), the LLM requires a decision-making mechanism to optimize resource efficiency. The goal is to minimize resource consumption, single tool calls can significantly reduce the LLM's interaction tokens. The experimental results in Table~\ref{tab:ablation_dpo} demonstrate that CMPO not only preserves Execution Agent's in-toolset selection performance but also substantially enhances out-toolset discrimination capability.

\begin{table}[h]
\scriptsize
\centering
\caption{Impact of CMPO on Execution Agent.}
\label{tab:ablation_dpo}
\resizebox{\linewidth}{!}{
\begin{tabular}{l*{4}{c}}
\toprule
\multicolumn{1}{c}{\textbf{Training Method}} & Shortest Path & Cycle Detection & Max Flow & Link Prediction \\
\midrule
Execution Agent w/o CMPO & 97.2 & 97.8 & 70.2 & 64.8 \\
Execution Agent w/ CMPO  & 98.2 & 99.0 & 89.7 & 85.2 \\
\bottomrule
\end{tabular}
}
\end{table}

\textbf{RQ5: How does Tool Creation Fine-tuning improve Tool Creator’s generation capability?} 

We select four out-toolset tasks within Graph4real on middle-scale. Three versions of the Tool Creator are evaluated: the original untuned agent (Llama3.1-8B), agent fine-tuned using Code-Enhanced SFT method, and agent fine-tuned using Tool Creation Fine-tuning method. Table \ref{tab:model_sft} reports the accuracy and executable percentage. The improvements in both accuracy and executability metrics confirm the effectiveness of Tool Creation Fine-tuning. Compared to Code-Enhanced SFT, especially for mainstream open-source LLMs, it is more challenging to enable LLMs to master the reasoning capabilities for complex tasks through code training. The complexity of the tasks and the limitations of LLM capabilities hinder their generalization performance on complex tasks. In contrast, Tool Creation Fine-tuning effectively enhances accuracy and executability by allowing the agent to focus more on generating core task-specific tools.

\begin{table}[h]
\small
\centering
\caption{Impact of Tool Creation Fine-tuning.}
\label{tab:model_sft}
\resizebox{\linewidth}{!}{
\begin{tabular}{l*{6}{c}}
\toprule
 & \multicolumn{2}{c}{\textbf{Without SFT}} & \multicolumn{2}{c}{\textbf{With Code SFT}} & \multicolumn{2}{c}{\textbf{With Tool SFT}} \\
\cmidrule(lr){2-3} \cmidrule(lr){4-5} \cmidrule(lr){6-7}
\multicolumn{1}{c}{\textbf{Task}}& Acc.(\%) & Execut.(\%) & Acc.(\%) & Execut.(\%) & Acc.(\%) & Execut.(\%) \\
\midrule
Max Flow & 38.2 & 58.4 & 72.2 & 81.2 & 89.6 & 96.3 \\
Common Neighbors & 42.6 & 63.1 & 70.6 & 78.3 & 88.8 & 95.8 \\
Link Prediction & 39.8 & 61.6 & 58.6 & 65.2 & 85.2 & 88.4 \\
Node Classification & 41.2 & 59.8 & 60.8 & 64.7 & 89.6 & 89.9 \\
\bottomrule
\end{tabular}
}
\end{table}

\textbf{RQ6: Can GraphCogent deal with multi-step tool planning tasks?}

To further probe our framework’s generalization to multi-step tool planning with interpretable reasoning paths, we introduce three classic multi-step reasoning tasks in Web scenarios: (i) Crawl-trap detection, which first detects redirect loops and plans a route; (ii) k-core filtered ranking, which computes PageRank on the k-core subgraph; and (iii) Capacity-aware pathing, which gates path computation with a capacity threshold before finding a shortest path. Each workflow is instantiated on middle-scale Web subgraphs. We compare four system variants: Naive Llama3.1-8B, TPI-only, CMPO-only, and GraphCogent (TPI+CMPO). We report two planning metrics: PlanEM, the percentage of cases where the predicted tool sequence exactly matches the reasoning chain; and TraceAlign-F1, the F1 between tool steps explicitly stated in the agent’s thinking trace and the actually executed steps. As shown in Table~\ref{tab:rq6_compose}, GraphCogent achieves consistently higher PlanEM and TraceAlign-F1 across all workflows, indicating more reliable sequencing and better trace–execution agreement; compare with TPI-only and CMPO-only, margin calibration reduces boundary confusions.

\begin{table}[h]
\small
\centering
\caption{GraphCogent's generalization to multi-step tool planning tasks. Trace-F1 denotes TraceAlign-F1.}
\label{tab:rq6_compose}
\resizebox{\linewidth}{!}{
\begin{tabular}{l*{6}{c}}
\toprule
& \multicolumn{2}{c}{Crawl-trap detection} & \multicolumn{2}{c}{k-core filtered ranking} & \multicolumn{2}{c}{Capacity-aware pathing} \\
\cmidrule(lr){2-3} \cmidrule(lr){4-5} \cmidrule(lr){6-7}
\multicolumn{1}{c}{Method} & PlanEM (\%) & Trace-F1 (\%) & PlanEM (\%) & Trace-F1 (\%) & PlanEM (\%) & Trace-F1 (\%) \\
\midrule
Naive Llama3.1-8B & 60.1 & 57.3 & 55.1 & 51.0 & 57.6 & 54.5 \\
TPI-only & 72.4 & 69.1 & 68.9 & 63.7 & 70.3 & 66.5 \\
CMPO-only & 83.8 & 80.3 & 79.7 & 75.9 & 82.5 & 78.8 \\
GraphCogent & 93.5 & 90.2 & 91.6 & 86.8 & 92.7 & 88.5 \\
\bottomrule
\end{tabular}
}
\end{table}

\section{Edit distance and Cost for GEC}
\label{GEC}

\begin{algorithm}[H]
\caption{Graph Edit Distance Computation}
\label{alg:edge_edit_distance}
\begin{algorithmic}[1]
\Require Source graph $G=(V,E)$; Target graph $H=(V,E')$ with identical node set $V$
\Ensure Edit distance $d_{\text{edit}}(G,H)$
\State \textbf{Assert} $V(H)=V(G)$
\State $E_G \gets \operatorname{set}(E),\quad E_H \gets \operatorname{set}(E')$  
\If{$E_G = E_H$} \State \Return $0$ \EndIf
\State $\mathcal{A} \gets E_H \setminus E_G,\quad \mathcal{R} \gets E_G \setminus E_H$
\State $a \gets |\mathcal{A}|,\quad r \gets |\mathcal{R}|$
\State $w_{\text{add}} \gets 1,\quad w_{\text{rem}} \gets 1$
\State $d_{\text{edit}}(G,H) \gets w_{\text{add}}\cdot a + w_{\text{rem}}\cdot r$
\State \Return $d_{\text{edit}}(G,H)$
\end{algorithmic}
\end{algorithm}

\subsection{Edit distance}
\label{GECEditdistance}

Real-world scenarios present graphs in multiple text forms—adjacency lists, symbolic notations, and linguistic descriptions—making rule-based preprocessing (e.g., regular expressions) impractical for standardization. Following human external sensory processing, our Sensory Module converts these varied inputs into a standardized adjacency-list representation. Because small transformation errors can propagate downstream reasoning on topology-rich graphs, we perform rigorous quality assessment using edge edit distance (Algorithm~\ref{alg:edge_edit_distance}), which measures the minimum edge additions and deletions needed to transform one graph into another.

\subsection{Cost for Graph Efficiency Coefficient}
\label{GECLLM}

We select token count as GEC’s cost metric for three reasons: (i) token-based pricing is the universal billing standard across LLM platforms (GPT, Claude, DeepSeek), enabling fair comparison; (ii) in tests on 100 graphs (300 edges) with four granularities, processing time varied by less than 5\%, since transformation outputs are length-stable, so time is not discriminative; (iii) the dominant expense is input tokens from the fixed instruction used to standardize graphs. We therefore use normalized token counts to capture the granularity trade-off: finer sampling increases LLM calls and accumulates instruction tokens, whereas coarser sampling reduces calls at the risk of topological distortion.

\section{Error Analysis}
\label{Error Analysis}
In this section, we analyze the errors of GraphCogent within the Graph4real benchmark for both in-toolset and out-toolset tasks.

\subsection{In-toolset tasks}

We select four in-toolset tasks and analyze error patterns across 500 test instances at the middle scale (100 nodes). The errors are categorized into two types: (i) \textit{Edge Error}: Structural deviations between the original graph and the transformed graph caused by inconsistent edges during the transformation. (ii) \textit{Tool Error}: Incorrect tool selection or parameterization in the Execution Module.

\begin{table}[h]
\centering
\small
\caption{Error distribution across in-toolset tasks.}
\label{tab:error_analysis1}
\begin{tabular}{l|cc|c}
\toprule
\textbf{Task Type} & \textit{Edge} & \textit{Tool} & Accuracy (\%) \\
\midrule
Edge Count & 19 & 2 & 97.8 \\
Triangle Count & 18 & 1 & 98.8 \\
Cycle Detection & 21 & 5 & 99.0 \\
Path Existence & 19 & 13 & 97.4 \\
\bottomrule
\end{tabular}
\end{table}

The results in Table~\ref{tab:error_analysis1} demonstrate that GraphCogent achieves high accuracy across in-toolset tasks, but performance varies by task type. For edge-dependent tasks (e.g., Edge Count, Triangle Count), errors primarily stem from Edge Errors, structural discrepancies introduced during graph transformation. These tasks rely on precise edge information, making them sensitive to minor deviations in the adjacency list conversion. In contrast, topology-independent tasks (e.g., Cycle Detection, Path Existence) show a reversed pattern. While Edge Errors may occur, their impact is mitigated because task outcomes do not depend on specific edge subsets. Therefore, Tool Errors dominate the error types for these tasks.

\subsection{Out-toolset tasks}
For out-toolset tasks, we select Max Flow on Transportation and Link Prediction on Social, analyzing error patterns across 500 test instances at the middle-scale graph (100 nodes). We classify errors into three categories: (i) \textit{Discrimination Error}: Occurs when the Execution Agent fails to identify the task as an out-toolset task, leading to inappropriate tool calls instead of activating the Tool Creator. (ii) \textit{Execution Error}: Arises when the task-specific tool generated by the Tool Creator contains logical flaws or implementation issues that prevent successful execution. (iii) \textit{Format Error}: Arises when the task-specific tool produces correct results but returns them in non-compliant formats.


\begin{table}[h]
\centering
\small
\caption{Error distribution across out-toolset tasks.}
\label{tab:error_analysis2}
\small
\begin{tabular}{l|ccc|c}
\toprule
\textbf{Task Type} & \textit{Discrimination} & \textit{Execution} & \textit{Format} & Accuracy (\%) \\
\midrule
Max Flow & 35 & 6 & 11 & 89.6 \\
Link Prediction & 6 & 58 & 10 & 85.2 \\
\bottomrule
\end{tabular}
\end{table}

The results are reported in Table~\ref{tab:error_analysis2}. For Max Flow, discrimination errors dominate, primarily due to the Execution Agent's failure to discriminate task boundaries when problems exceed the toolset's coverage. In such cases, the Execution Agent incorrectly attempts tool-based solutions despite the task requiring tool creation. Format errors are secondary, often manifesting as mismatched outputs (e.g., returning paths instead of the max flow value). In contrast, Link Prediction is primarily affected by execution errors, which stem from internal tensor mismatches in the generated neural models (e.g., dimension inconsistencies in GCN layers or invalid adjacency matrix operations). Both discrimination and format errors occur less frequently, suggesting that the model reliably identifies the task type and produces valid outputs.

\section{GraphCogent Workflow}
\label{GraphCogent Workflow}
Algorithm~\ref{alg:graphcogent} presents the workflow of GraphCogent. More detailed workflow and corresponding prompt templates are provided in our released anonymous github code.

\begin{algorithm}[H]
\caption{GraphCogent Framework}
\label{alg:graphcogent}
\begin{algorithmic}[1]
\Require Graph task $T$, Input graph $G$
\Ensure Task result $R$

\State \textbf{Sensory Module:}
\If{$G$ is file-based (edgelist/txt/json)}
    \State Bypass transformation; forward $G$ directly to Buffer
\Else
    \State Apply sliding-window sampling (GEC-selected; default 50 edges with 5\% overlap)
    \State Transform text to adjacency lists via heuristic prompts
    \State Verify via Graph Verifier (quantity consistency + format compliance)
\EndIf

\State \textbf{Buffer Module:}
\State Construct complete graph from adjacency lists
\State Transform to multiple representations: NetworkX/PyG/NumPy
\State Build data indices (dimensionality, schema, metadata)
\State Store preprocessed graph data $\tilde{G}$

\State \textbf{Execution Module:}
\If{$T$ is in common toolset coverage}
    \State Select appropriate tool from common toolset
    \State Retrieve required data format from Buffer Module
    \State Execute tool calling on $\tilde{G}$; \Return $R$
\Else
    \State Generate task-specific model $M$ for $T$
    \State Combine $M$ with $\tilde{G}$ from Buffer Module
    \State Execute generated model; \Return $R$
\EndIf
\end{algorithmic}
\end{algorithm}





\end{document}